\documentclass[10pt,twocolumn,letterpaper]{article}

\usepackage[pagenumbers]{cvpr} 
\makeatletter
\@namedef{ver@everyshi.sty}{}
\makeatother

\usepackage{graphicx}
\usepackage{amsmath}
\usepackage{amssymb}
\usepackage{booktabs}
\usepackage[inline]{enumitem}

\usepackage[pagebackref,breaklinks,colorlinks]{hyperref}
\usepackage{svg}

\usepackage[capitalize]{cleveref}
\crefname{section}{Sec.}{Secs.}
\Crefname{section}{Section}{Sections}
\Crefname{table}{Table}{Tables}
\crefname{table}{Tab.}{Tabs.}

\def\cvprPaperID{9046} 
\def\confName{CVPR}
\def\confYear{2022}

\begin{document}

\title{Anchoring to Exemplars for Training Mixture-of-Expert Cell Embeddings}

\author{Siqi Wang\thanks{Equal contribution}\\
Boston University\\
{\tt\small siqiwang@bu.edu}
\and
Manyuan Lu\footnotemark[1]\\
Boston University\\
{\tt\small manyuan@bu.edu}
\and
Nikita Moshkov\\
Biological Research Centre\\
{\tt\small nikita.moshkov@brc.hu}
\and
Juan C. Caicedo\\
Broad Institute of MIT and Harvard\\
{\tt\small jcaicedo@broad.mit.edu}
\and
Bryan A. Plummer\\
Boston University\\
{\tt\small bplum@bu.edu}
}

\maketitle

\begin{abstract}
Analyzing the morphology of cells in microscopy images can provide insights into the mechanism of compounds or the function of genes. Addressing this task requires methods that can not only extract biological information from the images, but also ignore technical variations, \ie, changes in experimental procedure or differences between equipments used to collect microscopy images. We propose Treatment ExemplArs with Mixture-of-experts (TEAMs), an embedding learning approach that learns a set of experts that are specialized in capturing technical variations in our training set and then aggregates specialist's predictions at test time. Thus, TEAMs can learn powerful embeddings with less technical variation bias by minimizing the noise from every expert. To train our model, we leverage Treatment Exemplars that enable our approach to capture the distribution of the entire dataset in every minibatch while still fitting into GPU memory.  We evaluate our approach on three datasets for tasks like drug discovery, boosting performance on identifying the true mechanism of action of cell treatments by 5.5-11\% over the state-of-the-art.
\end{abstract}

\begin{figure}[t]
\centering
\includegraphics[width=0.48\textwidth]{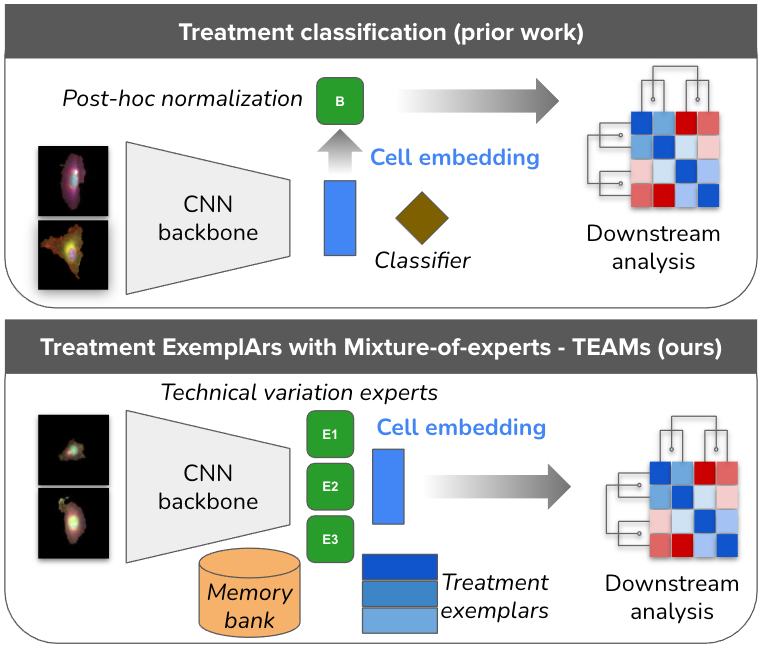}
\caption{\textbf{Strategies to compute single-cell embeddings.} \textbf{(top)} Treatment classification optimizes a CNN to identify biological treatments, which are weak labels with respect to the true mechanism of action \cite{Caicedo_2018_CVPR}. \textbf{(bottom)} Proposed model: a metric learning approach with a mixture of experts (to capture technical variation), treatment exemplars (to capture phenotype distributions), and a memory bank (to facilitate learning).}
\label{Fig:overview}
\end{figure}

\section{Introduction}

Cell images can be used to infer the effects and mechanisms of compounds in many contexts, following an approach known as image-based profiling \cite{pratapa2021image}, which requires a comprehensive and generic feature representation of cell morphology. This task has traditionally been addressed with hand-engineered features~\cite{rohbanCell2019}, and more recently, transfer learning with models pre-trained on natural images~\cite{ando2017improving,pawlowski2016automating}. Training models directly on cellular images holds the potential to improve the sensitivity of biological experiments \cite{chandrasekaran2021image,pratapa2021image}. However, ground truth annotations are not available for supervised training, primarily because the effects (or mechanisms) of compounds are usually not known in advance in real world conditions (\eg drug discovery). Thus, prior work has either used a small dataset with a limited set of ground-truth mechanistic classes \cite{godinez2017multi}, or, more commonly, used treatment labels for classification instead of ground-truth mechanistic classes~\cite{Caicedo_2018_CVPR}. Given that treatment labels are always known (\ie chosen by scientists), they can be used as weak labels to supervise a model with the expectation that the latent space captures relevant properties of mechanistic classes.

\begin{figure*}[t]
\centering
\includegraphics[width=\textwidth]{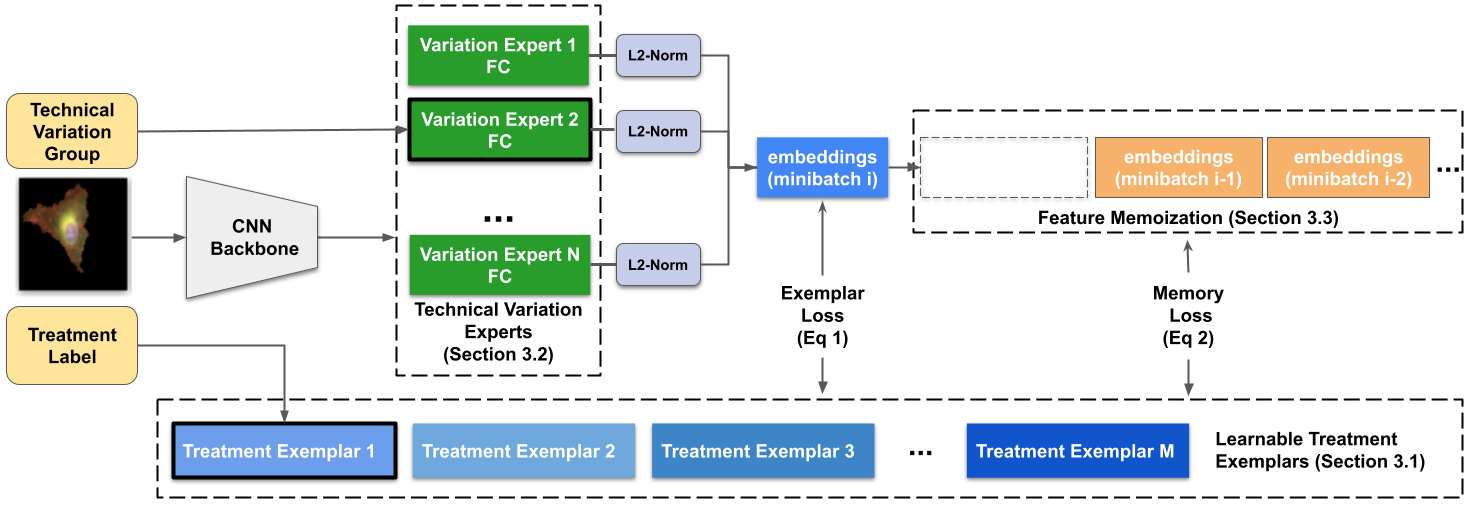}
\caption{\textbf{TEAMs during training.} The training input has three parts: a single-cell image, technical variation group, and treatment label. First, a base feature representation is generated from a CNN backbone (ResNet-18~\cite{he2016deep} in our experiments). Then it is transformed by an variation expert specified by the cell's technical variation group. Note that only one expert is activated according to the group input (although different experts may be used for other cells in a minibatch). Next, the cell embedding is compared with learnable Treatment Exemplars, which model the canonical representation of the cells for each treatment. Finally, feature memoization is used to efficiently increase batch size by reusing embeddings from recent minibatches. See Figure~\ref{fig:inference} for an illustration of using TEAMs for inference.}
\label{Fig:architecture}
\end{figure*}

While using (weak) treatment labels directly in classification models has shown to be useful, there are limitations in generalizing features that make it not optimal for image-based profiling. The power of image-based profiling is on modeling morphology as a continuous transition of cell states that can be compared, rather than as categorical, discrete groups. The categorical classification loss, typically used in weakly supervised learning, imposes a discrete grouping that may separate cells not by their relevant biological traits, but by other irrelevant factors of variation, including technical artifacts, which can affect the ability of these models to generalize.  As illustrated in Figure~\ref{Fig:overview}, prior work (\eg,~\cite{Caicedo_2018_CVPR}), has typically addressed this through a post-hoc normalization approach using cells from a control group, but such an approach is ineffective if informative features were not learned in the initial training step.

In this work, we propose Treatment ExemplArs with Mixture-of-experts (TEAMs), a metric learning approach for training single-cell representations from biological images.  Our approach has three components, each of which addresses a complementary challenge.  First,  learnable Treatment Exemplars transform the metric learning problem into a type of cluster prediction task, where single cells are encouraged to embed nearby the exemplars from the treatments that produced them.  This provides an efficient mechanism for capturing the distribution of the entire dataset in each minibatch, ensuring that informative samples are always present.  
Second, we use variation experts to learn embeddings that are specialized to a specific set of technical variations. Each projection learned by the expert takes cell features from some shared general embedding space to a new subspace. Then, at test time we mix together the predictions of all of our experts.  This helps us avoid overfitting to just one setting, reducing variance due to spurious correlations caused by a single set of variations.  Finally, we use a cross-batch memory module~\cite{wang2020xbm} that reuses samples computed in recent batches to efficiently increase the batch size, boosting the information used to update model parameters.  Figure~\ref{Fig:architecture} contains an overview of our approach.

To summarize, our contributions are:
\begin{itemize}
\setlength\itemsep{0em}
    \item We introduce TEAMs, a novel feature learning approach that improves performance on downstream tasks like identifying the true mechanism of action of cell treatments by 5.5-11\% over the state-of-the-art. 
    \item We show that our Mixture-of-Expert cell embeddings have less technical variation defects. Compared with baselines like adversarial alignment~\cite{ganinICML2015} or self-challenging~\cite{huangRSC2020}, our method outperforms by 1-1.5\%.
    \item We demonstrate that our learnable Treatment Exemplars can boost downstream performance by learning more informative features due to seeing the entire distribution of the training data in each minibatch.
\end{itemize}

\section{Related Work}
\noindent\textbf{Deep learning for image-based profiling.} Cell morphology has traditionally been measured with classic hand-crafted features in high-throughput biological experiments \cite{caicedo2017data}. In early work convolutional networks demonstrated how models pre-trained on ImageNet can be used to obtain cell morphology embeddings \cite{pawlowski2016automating, ando2017improving, godec2019democratized, jackson2019phenotypic}. This transfer learning approach can improve performance without requiring any training: only a post-hoc re-normalization is applied to calibrate the relevant factors of variation.

Training feature extraction models directly on cell images is attractive given the large amounts of data acquired in high-throughput experiments (thousands of compounds are tested in parallel). However, well curated and manually annotated datasets with ground truth labels of cellular phenotypes do not exist. Several studies have trained neural networks with cellular images using unsupervised learning, including variational auto-encoders \cite{lafarge2019capturing}, image in-painting \cite{lu2019learning}, deep clustering \cite{janssens2020fully}, and contrastive learning \cite{perakis2021contrastive}. 

Weakly supervised learning was proposed to train models using treatment labels as a pretext task \cite{Caicedo_2018_CVPR}. Treatment labels are always known for a given high-throughput experiment, but they are considered weak labels because the true phenotypic response of cells (also known as mechanism of action) is not known for all treatments. The approach, like the other methods we have discussed, may require post-hoc feature re-normalization as well to account for technical variation.  That said, it has been successfully adopted to analyze treatments for diseases such as COVID-19 \cite{cuccarese2020functional}, and anti-aging treatments \cite{white2020multi}.  In contrast, our TEAMs model is designed to minimize the effect of technical variation without requiring any post-hoc processing steps.
\smallskip

\noindent\textbf{Deep metric learning.} Most relevant to our work are methods that learn multiple embedding spaces within a single model (\eg,~\cite{LIN_2020_CVPR,mishraPAN2021,plummer2018conditional,plummerHint2019,tan2019learning,vasileva2018learning,veit2017conditional}).  However, most models have been developed for different settings where they either assume they know something about what embedding to use at test time, whereas for our task we make no such assumptions.  In contrast, we use our embedding spaces so that our model becomes an expert at capturing information about cells where they have similar technical variations, then minimize the effect of spurious correlations by aggregating predictions across experts.  This goal is similar to methods that were developed to generalize across domains (\eg~\cite{huangRSC2020,kimMULEAAAI2020,Saito_2018_CVPR}), but this work typically involves trying to align feature distributions of different domains into a shared embedding space.  However, knowing what constitutes an artifact rather than an informative feature is not known beforehand, so methods focusing on alignment may unintentionally remove informative features, whereas our approach can still take advantage of them with our experts.


\section{Treatment ExemplArs with Mixture-of-experts (TEAMs)}

Given two images of single cells, our goal is to measure if they share a mechanistic class. More formally, given a pair of images $(I_a, I_p)$ our goal is to embed them nearby each other if they were from the same mechanistic class, \ie, $M_a = M_p$, or to embed them far away from each other otherwise, $M_a \neq M_p$. However, since mechanistic classes are not available during training, we use the type of treatment given to a cell as a proxy label for mechanistic classes. Then, $I_a, I_p$ are embedded according to treatment labels $T_a$ and $T_p$. A standard way of learning this kind of embedding is using a triplet loss, but performance is often very sensitive on the size of minibatches used for training and how the images are sampled~\cite{wu2017sampling}.  In addition, technical variations may cause a distribution mismatch between training and inference~\cite{Caicedo_2018_CVPR}, reducing performance.  To address these issues, we introduce TEAMs, which has separate modules used to address each problem we outlined above.  Specifically, Section~\ref{subsec:proxy} transforms our representation learning task into a cluster prediction task that effectively allows us to represent the entire distribution of our training in every minibatch data without requiring expensive offline sampling techniques, Section~\ref{subsec:independ} learns ensembles of experts that minimizes the effect of technical variations, and Section~\ref{subsec:memory} improves the quality of our gradients by effectively increasing the minibatch size using a memory module.

\subsection{Training with Learnable Treatment Exemplars}
\label{subsec:proxy}

Training with the entire dataset in a single minibatch is often not possible due to limitations in GPU memory.  Thus, minibatches are employed that typically represent a very small subset of the data.  Selecting informative samples can have a significant impact on performance~\cite{wu2017sampling}. 
Thus, researchers using several intelligent sampling techniques often focus on finding hard negatives (\eg~\cite{Schroff_2015_CVPR,Wang_2016_CVPR}), but this effectively introduces a bias during training due to focusing more on certain samples.
Wu~\etal~\cite{wu2017sampling} demonstrated that having a sampling method that is more representative of the distribution of the dataset can improve performance.  
However, they relied on an expensive offline sampling approach, where pairwise comparisons between all training samples are needed. The statistics about the distribution of the dataset are collected every few epochs during training, which can be prohibitively expensive for very large datasets.

Inspired by recent work in metric learning~\cite{Lu_2021_CVPR,Movshovitz-Attias_2017_ICCV,tehProxy2020}, we reformulate our problem such that each image is compared with an exemplar for every treatment in our training set.  These treatment exemplars are learnable parameters that are meant to represent the canonical representation of a cell that underwent a specific treatment in our training set.  Since these exemplars are the same size as the final representations of our cells (512-D in our experiments) they are extremely memory efficient.  As a result, in our experiments we are able to compare every image in a minibatch with the exemplars for treatments that represent the entire dataset, which contains images of over a million cells, thereby accurately representing the entire dataset distribution in a minibatch.  Our exemplars are trained by predicting the treatment that produced each cell.  Given treatment $t \in T$ and a corresponding $\ell_2$-normalized image features of a cell $I_t$, and a set of $\ell_2$-normalized exemplars $C$ for each treatment (\ie, $|C| = |T|$), we minimize the cosine distance $d(\cdot)$ between image $I_t$ and its corresponding exemplar $C_t$ while maximizing its distance to all other exemplars $Z$, \ie:

\begin{equation}
    L_{Exemplar} = -\log \left(\frac{\exp(-d(I_x, C_x))}{\sum_{y \in Z} \exp(-d(I_x, C_y))}\right).
    \label{eq:proxy}
\end{equation}

Note that we also explored modifications such as giving exemplars a higher learning rate or using temperature scaling that was shown to be beneficial in prior work~\cite{tehProxy2020}, but found they did not improve performance in our experiments.

\subsection{Minimizing Noise with Mixtures of Technical Variation Experts}
\label{subsec:independ} 
Technical variations in how cell images were collected can create a shift in the distribution between datasets. Working on images from different distributed datasets can be turned into the problem of aligning features across domains. \eg, using an adversarial classifier~\cite{ganinICML2015} to solve the mismatches between train/test distributions.  Self-challenging has also been used to create models that are more robust across domains by ensuring many different kinds of features are learned~\cite{huangRSC2020}.  This suggests that learning multiple kinds of features, or, in other words, different specialized experts, can help improve performance across domains. However, Huang~\etal~\cite{huangRSC2020} learned these experts by forcing non-dominant features to activate according to the labels, but doesn't explicitly model the different distributions that can occur across different domains, which limits its expressiveness.  
In contrast, we use a mixture-of-experts approach that ensembles a set of variation-specific experts learning different distributions separately to minimize the technical noise.

Let $V$ be a set of training variations, where each variation is a certain technical setting to collect the data. For each variation, we train a variation-specific expert to learn the projection $W$, which transforms the cell features from a shared network backbone to the final variation-specific representation. The projection $W$ is supposed to minimize the specific technical variance noise and only output informative biological features as the variation-specific representation. Then the cell images from different variations can be compared with the single set of Treatment Exemplars in Section~\ref{subsec:proxy}, \ie, we compute $I'_x = W_v f(I_x)$ for Eq.~(\ref{eq:proxy}), where $v \in V$ and $f(I_x)$ represents the cell feature from our shared network backbone. In our experiments, an 18-layer deep residual network~\cite{he2016deep} is used, but our approach generalizes to any architecture (see supplementary for EfficientNet-B0~\cite{efficientnet} results). Using a common network backbone encourages feature sharing across settings while also minimizing the amount of specialized knowledge that can be captured by using a single linear projection, which can reduce overfitting to a single set of technical variations.

\begin{figure}
\centering
\includegraphics[width=0.48\textwidth]{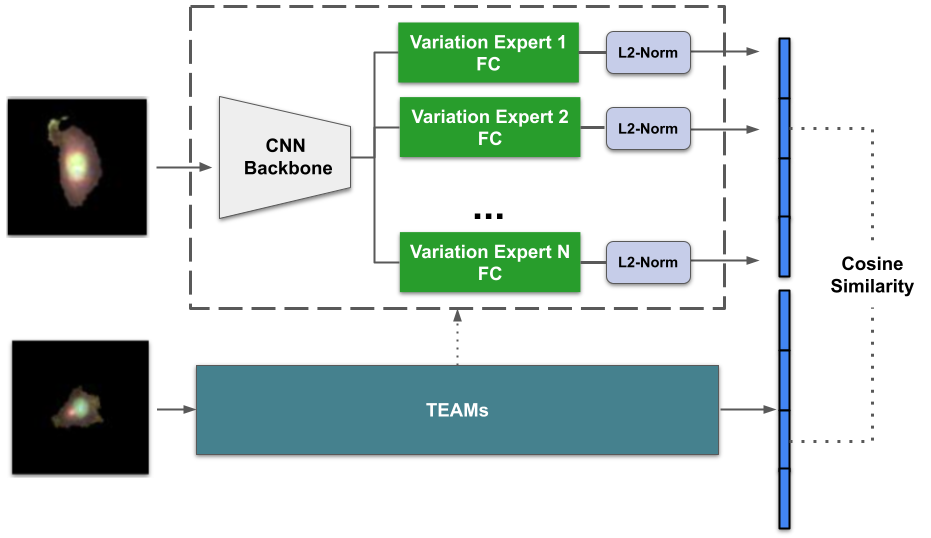}
\caption{\textbf{TEAMs during inference.} During training, technical variation group is know to us, and hence can choose expert to use for each cell image.  However, at test time we assume we are not provided with this information, and selecting a single expert may also result in a biased prediction.  Thus, we found that averaging the predictions of all experts by concatenating the features together results in best performance.}
\label{fig:inference}
\end{figure}

During inference, however, little information about the variation distribution is known. Selecting a single variation-specific expert could be challenging, and no single expert may accurately represent the new data.  A single expert may also contain spurious correlations from its observed technical variations.  Instead, we represent a cell image by obtaining the variation-specific representation from each of our experts and concatenating them together (illustrated in Figure~\ref{fig:inference}).  Since we use a shared network representation, obtaining features from all our experts is efficient since are all obtained using a single linear projection.  Note that we $\ell_2$-normalize each variation-specific representation separately before concatenation so that they accurately represent the features learned during training.  Our mixtures-of-experts approach is similar to Wang~\etal~\cite{wang2020fair}, which addressed the task of promoting fairness in image classification, where the goal is to minimize the performance differences across a set of protected attributes for a task.  Our mixtures-of-experts can be seen as an adaptation of their method to our task of learning robust representations of cell images.

\subsection{Increasing Minibatch Information with Feature Memoization}
\label{subsec:memory}

\begin{figure}
\centering
\includegraphics[width=0.5\textwidth]{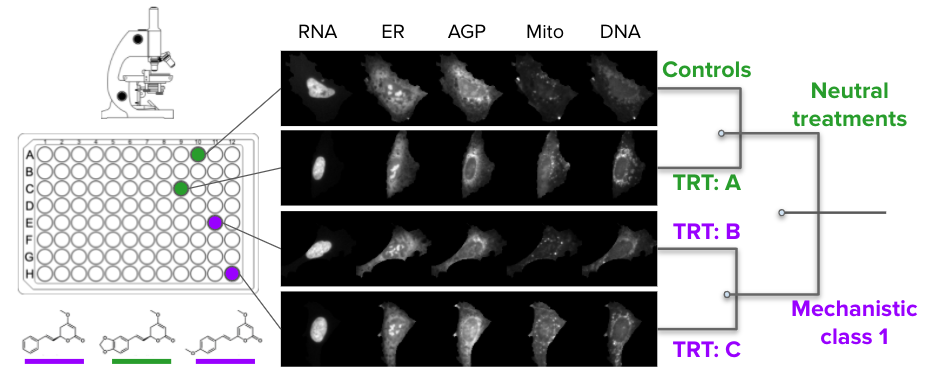}
\caption{\textbf{Cell Painting Example.} Images are from three treatments and one control, sampled from a high-throughput experiment with thousands of treatments. The morphology of cells reveals their response to treatments, which can be used to identify effectiveness and relationships with other treatments, according to mechanism of action classes.}
\label{Fig:task}
\end{figure}

\begin{table*}[th]
    \centering
    \setlength{\tabcolsep}{2.5pt}
    \begin{tabular}{|c|c|c|c|c|}
    \hline
     Experiment & Anchor & P & N & notations\\
    \hline
    1 & $I_a \in m$ & $I_p \in m$ & $I_n \not\in m$ & $m \in M_{test}$ \\
    2 & $I_a \in m$ & $I_p \in m$ & $I_n \in$ control group & $m \in M_{test}$ \\
    3 & $t_a \in m$ & $t_p \in m$ & $t_n \not\in m$& $m \in M_{test}$, $t_a, t_p, t_n \in T_{test}$\\
    \hline
    \end{tabular}
    \caption{\textbf{Experiments Metrics}. $I$ is a single cell image; $m$ is a mechanistic class; $M_{test}$ is a set of mechanistic classes in test dataset; $t$ is a treatment, which is a set of images here: $t_k = \{I|I_{treatment} = k\}$; $T_{test}$ represents the treatments in test dataset. }
    \label{tab:exp_metrics}
\end{table*}

Our Treatment Exemplars described in Section~\ref{subsec:proxy} provide a mechanism where each cell image is compared with the entire distribution of the dataset in a minibatch.  However, the converse is not true, each exemplar may not have any representative samples from the treatment it represents, and instead is only updated with information from negative samples.  If we alter our minibatches to have a balanced number of cells from each treatment, they may still not fit into GPU memory in datasets with many treatments.  Wang~\etal~\cite{wang2020xbm} observed that embeddings between minibatches in close proximity during training do not change much, and would likely provide a very similar signal if they were simply retained in the nearby subsequent training iterations rather than recomputed.   Thus, we keep the $K$ most recent embeddings (but not their underlying gradients) in memory module $\Phi$ and simply reuse them in future minibatches to compare them with their proxies via:
\begin{equation}
    L_{memory} = -\log \left(\frac{\exp(-d(\Phi_x, C_x))}{\sum_{y \in Z} \exp(-d(\Phi_x, C_y))}\right).
    \label{eq:memory}
\end{equation}
The number of iterations a sample is retained is dependent on the relative size of $K$ and the minibatch size.  For example, if $K=256$ and the minibatch has 128 samples, then each sample would be retained for $256/128=2$ iterations. The total loss function for TEAMs is a linear combination of our memory and exemplar losses, \ie, 
\begin{equation}
    L_{total} = L_{Exemplar} + L_{memory}
\end{equation}
By effectively increasing the minibatch size using our memory module, each exemplar is more likely to be provided with samples from its own treatment during training, making gradient updates more effective.

\section{Cell Painting Datasets}

Cell images for our experiments are from three Cell Painting \cite{bray2016cell} datasets, which represent large treatment screens of chemical and genetic perturbations. The BBBC022 and BBBC036 datasets are high-throughput compound screens testing 1,600 and 2,200 bioactive compounds, respectively. The BBBC037 dataset is a genetic over-expression screen of 200 genes. All three datasets\footnote{Available in the Broad Bioimage Benchmark Collection (BBBC) \url{https://bbbc.broadinstitute.org/image_sets}} were obtained by exposing U2OS cells (human bone osteosarcoma) to the treatments. Each treatment is tested in 5 replicates, using multi-well plates, and then imaged with the Cell Painting protocol \cite{bray2016cell}, which is based on six fluorescent markers captured in five channels (Figure \ref{Fig:task}). 
\smallskip

The cell population in each replicate is captured with up to 9 images (typically 1080$\times$1080 pixels) using the same magnification (20X). One image may contain hundreds of cells, and we identify each one to quantify treatment effects at single-cell resolution. Our goal is to model cell morphology features that capture similarities between the effects of treatments, which we evaluate by looking into the ground truth \emph{mechanism of action classes}. The numbers of mechanistic classes in our datsets are 453,  693 and 29 for BBBC022, BBBC036, and BBBC037 respectively. 
\smallskip

\noindent\textbf{Data Pre-processing} All five-channel images were first processed with a retrospective illumination correction algorithm to fix uneven illumination distribution under the microscope objective \cite{singh2014pipeline}. Next, we run CellProfiler \cite{mcquin2018cellprofiler} to obtain cell segmentations using the seeded-watershed algorithm \cite{wahlby2004combining}. The locations of cell centers were recorded and used to crop-out individual cells in a 128$\times$128 image with five channels. We save each single cell in a separate 8-bit PNG file, with all channels concatenated in a single sequential strip (illustrated in Figure \ref{Fig:task}). 

\begin{table*}[t]
    \centering
    \setlength{\tabcolsep}{1.5pt}
    \begin{tabular}{|rl|ccc|c||ccc|c|}
    \hline
    & & \multicolumn{4}{|c||}{separate cell treatments} & \multicolumn{4}{|c|}{separate from controls}\\
    \cline{3-10}
     & & BBBC037 & BBBC036 & BBBC022 & Average & BBBC037 & BBBC036 & BBBC022 & Average\\
     \hline
    \textbf{(a)} & Transfer learning~\cite{efficientnet} &  51.7 & 53.2 & 52.6 & 52.5 & 52.2 & 76.3 & 71.5 & 66.6\\
        & Treatment classification~\cite{Caicedo_2018_CVPR} &  52.7 & 58.6 & 56.6 & 56.0 & 57.2 & 22.0 & 62.2 & 47.2\\
        \hline
    \textbf{(b)} & Online Negatives & 56.9 & 62.3 & 59.2 & 59.5 & 59.4 & 80.6 & 71.9 & 70.6\\
        & +Adversarial~\cite{ganinICML2015} & 57.8 & 63.4 & 60.5 & 60.6 & 59.6 & 80.8 & 70.1 & 70.2\\
        & +Self-Challenge~\cite{huangRSC2020} & 55.4 & 60.1 & 58.0 & 57.8 & 57.7 & 80.6 & 70.7 & 69.7\\
        & +Mixture-of-Experts (MoE) & 59.1 & 63.7 & 60.5 & 61.1 & 61.7 & 80.7 & 71.0 & 71.2\\
        \hline
    \textbf{(c)} & Exemplars & 57.8 & 64.3 & 62.7 & 61.6 & 61.1 & 73.3 & 70.8 & 68.4\\
        & Exemplars+MoE & 60.1 & 64.5 & 62.2 & 62.2 & 63.2 & 79.1 & 70.5 & 71.0\\
        & Exemplars+Memory & 57.8 & \textbf{64.9} & \textbf{62.6} & 61.8 & 61.0 & 73.5 & 70.9 & 68.4\\
        & TEAMs (Exemplars+MoE+Memory) & \textbf{60.5} & 64.7 & \textbf{62.6} & \textbf{62.6} & \textbf{63.6} & \textbf{81.5} & \textbf{72.4} & \textbf{72.5}\\
        \hline
    \end{tabular}
    \caption{\textbf{Single cell results.}  We compare how often a pair of images that share a mechanistic class are predicted as more similar than a pair of images that share no mechanistic class. \textbf{(a)} reports adaptations of prior work, \textbf{(b)} compares different ways of accounting for technical variation on top of our own strong baseline approach, and \textbf{(c)} contains ablations of our model.  See Section~\ref{subsec:single_cell} for discussion.}
    \label{tab:main_results}
\end{table*}
\subsection{Cell Representation Evaluation Protocols}
 For each pre-processed Cell Painting dataset, we perform 5-fold cross-validation where we split the images by treatment, ignoring images belonging to the controls group during training following~\cite{Caicedo_2018_CVPR}. For BBBC037 we use 108/30/10 treatments for train/test/validation resulting in an average of about 250K/75K/25K images per split, respectively. For BBBC036 we use 1151/300/100 treatments for train/test/validation resulting in an average of about 740K/200K/65K images per split, respectively. For BBBC022 we use 645/300/100 treatments for train/test/validation resulting in an average of about 470K/210K/73K images per split, respectively.
 
We evaluate performance across three experiments summarized in Table~\ref{tab:exp_metrics}. First, for every (anchor) image $I_a$ in the test (or validation) split, we sample two additional images $I_p, I_n$ to create a triplet of images. One of the sampled images is a cell image shares a mechanistic class $m$ with the anchor, whereas the other sampled image shares no mechanistic classes.  Both images are sampled at random from among the set that satisfies the constraints for that group.  Performance is measured by how often the image that shares a mechanistic class with the anchor is predicted as more similar than the image that shares no mechanistic classes.  As a reminder, the mechanistic class labels are not used during training, so this experiment is a type of a transfer learning problem.
The second experiment is similar, but the image that does not share a mechanistic class with the anchor is randomly selected from among the images in the control group.
The third experiment is analogous to the first, but the triplet being selected is over mechanistic classes. Specifically, for each (anchor) treatment $t_a$ in the test set, we select a pair of treatments $t_p, t_n$ where one shares a mechanistic class with the anchor, and the other treatment shares no mechanistic class. For treatment triplets, the similarity is computed by averaging pairwise similarity between all image pairs belonging to each treatment. For each split we randomly sample 1,000 treatment triplets for BBBC037 and 50,000 triplets for BBBC022 and BBBC036. 
 


\section{Experiments}

\noindent\textbf{Implementation details.}  We train TEAMs to create a 512-D embedding using Adam~\cite{DiederikICLR2015} with a learning rate of $1e^{-3}$, which is decayed exponentially using a gamma of 0.9.  We train for 40 epochs with a batch size of 768, using a memory size of 2,816 and select the epoch that performs the best on the validation set which is evaluated after every epoch.  We train our models using a single NVIDIA RTX A6000 GPU. Each Cell Painting dataset is treated as a set of technical variations (\ie, they each get their own experts).

\subsection{Compared Baselines}

\noindent\textbf{Transfer learning.} 
This approach uses a network pretrained on ImageNet as feature extractor of the morphology of the cell. First, each gray-scale channel of the cell is transformed into an RGB tensor by replicating its content in the channel axis. Then, each channel is processed separately by the pretrained network and their feature vectors are concatenated in a single representation. We used EfficientNetB0 ~\cite{efficientnet} in our experiments, and keep the features from the pooling layer before the classifier, resulting in a 6,400-D vector per cell (1,280 features per channel).
\smallskip 

\noindent\textbf{Treatment classification~\cite{Caicedo_2018_CVPR}.}
This approach follows a weakly supervised strategy to representation learning where treatment labels are used to train a classification network. The network is assigned the problem of classifying single cells into one of the known treatments, assuming that all the cells in a treatment display a similar phenotypic response. In our experiments, we trained an EfficientNetB0 backbone with one classifier for each dataset. 
We discard the classifier and keep features from the last pooling layer before the classifier, resulting in 1,280-D vectors.
\smallskip

\noindent\textbf{Online Negatives.}  We also compare to a model trained with a margin-based loss with online negative mining.  We assume we are provided with two pairs of images $(I_x, I_p)$ and $(I_y, I_n)$ which represent cells from the same and different treatments, respectively.  Note that unlike traditional triplet loss, where $I_x = I_y$, we found we could improve performance by relaxing this constraint by allowing the case where $I_x \neq I_y$, and computing the loss over all possible pairs within a minibatch.  \Ie, we sample minibatches by sampling pairs of cell images from the same treatment.  Then, we obtain their embeddings using the same encoder as our approach (ResNet-18~\cite{he2016deep}) and $\ell_2$-normalize them.  We then compute cosine similarity between all possible pairs of cell images in the minibatch and separate them into a set of positive image pairs $P$ for images from the same treatment and negative image pairs $N$ for images from different treatments.  Thus, our loss is computed as:
\begin{equation}
    H(I_x, I_p, I_y, I_n) = \max(0, m + d(I_y, I_n) - d(I_x, I_p))
\end{equation}

\begin{equation}
    L_{triplet} = \frac{1}{|P||N|}\\
    \sum_{\forall (I_x, I_p) \in P}\sum_{\forall (I_y, I_n) \in N}  H(I_x, I_p, I_y, I_n).
    \label{eq:triplet}
\end{equation}
\noindent where we set the margin $m=0.3$.
\smallskip

\noindent\textbf{Adversarial~\cite{ganinICML2015}.} This approach uses an adversarial classifier to align features across different domains.  This is implemented as a linear classifier that takes the features from our network backbone (\ie, those that are input into Eq.~(\ref{eq:triplet})) and then predicts the variation that the treatment came from.  Then, the gradients from the linear classifier that backpropagated into the network backbone from the adversarial classifier are flipped and scaled in order to encourage the underlying CNN to learn features that can't be used to discriminate between treatments.  We found a scaling factor of $1e^{-2}$ worked well in our experiments.
\smallskip

\noindent\textbf{Self-Challenge~\cite{huangRSC2020}.} This method develops more robust models by iteratively muting a percentage of the the most important features for making a prediction during training, thereby forcing the model to learn how to align additional features to the target labels. In our experiments, we set the percentage of features dropped at 50\% which is applied to 1/3rd of a minibatch. 
\smallskip

\subsection{Single-Cell Evaluation Results}
\label{subsec:single_cell}

Table~\ref{tab:main_results} reports our results separating single cells by mechanistic class as well as from controls.  Comparing our online negatives baseline in first line of Table~\ref{tab:main_results}(b) to prior work in Table~\ref{tab:main_results}(a) we see that our baseline provides a boost of 3.5-4\% on average.  While we show that using prior work for ignoring technical variation via methods like adversarial alignment can improve performance, we see that our Mixture-of-Experts (MoE) obtains best performance in Table~\ref{tab:main_results}(b).  We also report that our Treatment Exemplars reported in the first line of Table~\ref{tab:main_results}(c) give a 2\% boost when separating by mechanistic class over our online negatives baseline.  Note, however, this comes at a small cost of separating from controls.  This may be due to the fact that we do not explicitly model controls in our model, which may also be useful in identifying cells that have no reaction to their treatment.  We believe investigating this further may be a useful direction for future work.  Finally, we see through the ablations of our model in Table~\ref{tab:main_results}(c)  that each of the three components of TEAMs all provide meaningful contributions to obtain our best average performance across the three datasets, where we outperform prior work in Table~\ref{tab:main_results}(a) by 6-6.5\%.

Table~\ref{tab:one_out} reports the performance of training on only two of the three datasets. During training, only two variation experts are trained for the two datasets respectively. The results on the held-out dataset (diagonal values in the table) show how the experts work together for the new technical variation. For each column, comparing the results from the first three rows to the last row shows how the held-out dataset variation could help training. Notably, performance on the BBBC036 dataset is most improved when combined with all three datasets, while not training on BBC022 results in a significant drop in performance on that dataset. Comparing these results to the baselines in Table~\ref{tab:main_results}(a), we see that even when fewer images and examples of technical variation are available for training, our approach still provides up to a 3\% boost in overall performance.

\begin{table}
    \centering
    \setlength{\tabcolsep}{2.5pt}
    \begin{tabular}{|c|ccc|c|}
    \hline
     Held-Out & BBBC037 & BBBC036 & BBBC022 & Average \\
    \hline
    BBBC037 & \textbf{52.8} & 64.2 & 61.2 & \textbf{59.4}\\
    BBBC036 & 52.7 & 53.1 & \textbf{63.3} & 56.4\\
    BBBC022 & 52.4 & \textbf{65.2} & 53.3 & 57.0\\
    \hline
    None & 60.5 & 64.7 & 62.6 & 62.6\\
    \hline
    \end{tabular}
    \caption{Effect of fewer examples of technical variation has on single cell performance using TEAMs.  We see that with more examples of technical variation we obtain better average performance across all settings.  See Section~\ref{subsec:single_cell} for additional discussion.}
    \label{tab:one_out}
\end{table}

\begin{table*}
    \centering
    \setlength{\tabcolsep}{2.pt}
    \begin{tabular}{|l|ccc|c|}
    \hline
     & BBBC037 & BBBC036 & BBBC022 & Average \\
    \hline
    Transfer learning~\cite{efficientnet} & 70.2 & 58.9 & 54.3 & 61.1 \\
    Treatment classification~\cite{Caicedo_2018_CVPR} &  81.2 & 55.6 & 53.3 & 66.7\\
    TEAMs & \textbf{84.8} & \textbf{67.4} & \textbf{64.2} & \textbf{72.1} \\
    \hline
    \end{tabular}
    \caption{\textbf{Treatment-level results.}  We compare how often a method can accurately identify which treatments share a mechanistic class after averaging the pairwise similarity scores between all unique pairs of cells from different treatments.  See Section~\ref{subsec:treatment} for discussion.}
    \label{tab:treatment}
\end{table*}

\begin{figure*}[t]
\centering
\includegraphics[width=\textwidth]{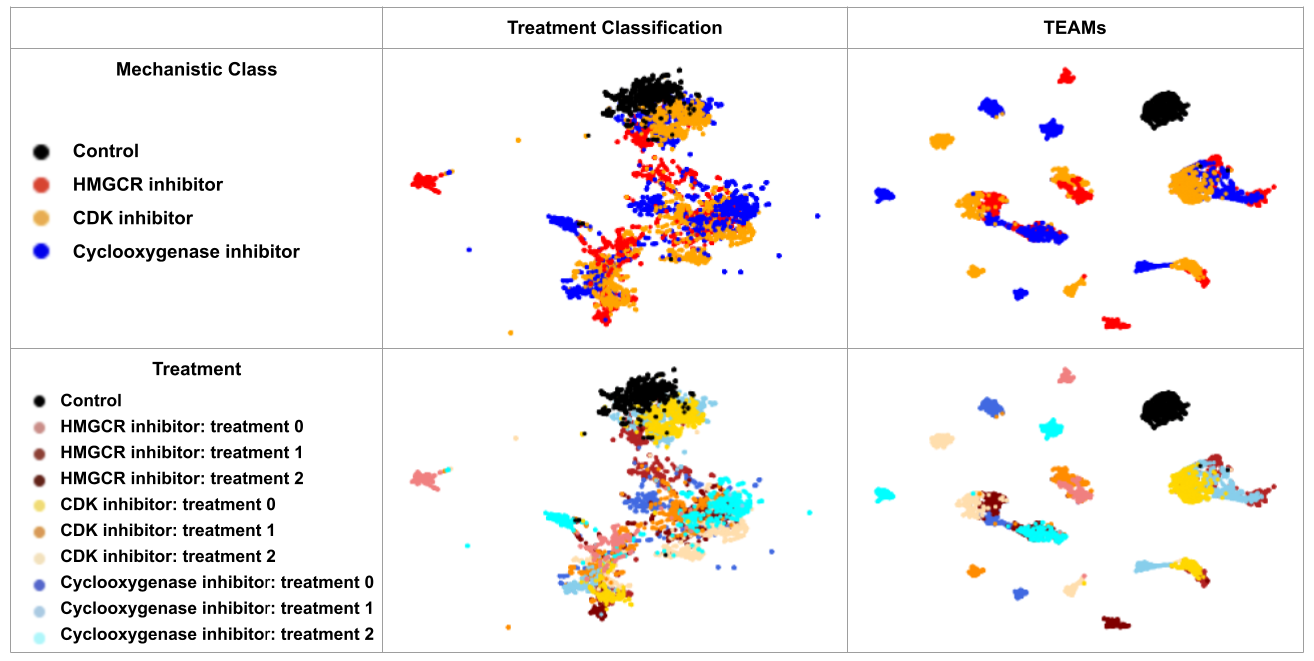}
\caption{\textbf{Cell-embedding visualization with UMAP.} For each mechanistic class, three treatments are selected to show finer clusters. When contrasting our approach with Treatment Classification~\cite{Caicedo_2018_CVPR}, it is clear that TEAMs has well separated clusters with far less noise compared to the baseline. See Section~\ref{subsec:treatment} for additional discussion.}
\label{Fig:qualitative results}
\end{figure*}

\subsection{Treatment-Level Results}
\label{subsec:treatment}

Table~\ref{tab:treatment} shows how TEAMs compares to prior work when aggregating predictions over entire cell treatments in order to identify treatments that share mechanistic classes.  Overall TEAMs boost performance between 5.5-11\% over prior work across the three datasets.  Over individual datasets, the largest boost in performance is seen over the BBBC036 dataset, which improves over the best model from prior work by 8.5\%.
\smallskip


\noindent\textbf{Qualitative results.}  Figure~\ref{Fig:qualitative results} compares the embeddings learned by our approach visualized with UMAP~\cite{mcinnes2018umap-software} to those created by Treatment Classification~\cite{Caicedo_2018_CVPR}.  We see in Figure~\ref{Fig:qualitative results}(top) that TEAMs is able to create more homogeneous clusters compared to Treatment Classification, although this includes creating multiple clusters of cells that share mechanistic classes.  We note that some treatments that are labeled as having different mechanistic classes may actually have some undiscovered similarities in mechanistic classes as well.  However, as can be seen in Figure~\ref{Fig:qualitative results}(bottom), much of the separation in the same mechanistic class can be accounted for by coloring the clusters by treatment.  This suggests that TEAMs is far more capable at learning a good embedding for the training task, but that the features learned do not always translate perfectly to the target task.  In addition, there may be some effect due to technical variation that could be further improved in future work.  Our quantitative results still demonstrate that despite these limitations our approach is more effective than prior work.

\subsection{Ethical considerations}

This study was conducted with biological images of human bone osteosarcoma cells, an immortalized cell line used for research purposes only. The images or data in this study do not contain patient information of any kind. The use of these images, and the algorithms to analyze them, is to test the effects of treatments. Automating drug discovery has positive impacts in society, specifically the potential to help finding cures for diseases of pressing need around the world in shorter times, and utilizing less resources. The proposed methods could be used to optimize drugs that harm people; we do not intent that as an application, and we expect regulations in biological labs to prevent such uses.

\section{Conclusion}
We provide a novel metric learning approach: Treatment ExemplArs with Mixture-of-experts (TEAMs) to capture biological features from single-cell images. Learnable treatment exemplars, technical variation experts and feature memoization are three key modules of the model. Learnable treatment exemplars represent the entire distribution of the dataset in every minibatch data, variation experts minimize the noise from technical artifacts and memory module effectively increases the minibatch size without extra computation. TEAMs outperforms the state-of-the-art by 5.5-11\% on downstream tasks like identifying the true mechanism of action of cell treatments. Identifying cells that have no reaction to treatment and translating features from training task to test task could be directions for future work.

{\small
\bibliographystyle{ieee_fullname}
\bibliography{egbib}
}
\appendix
\section{Comparing Network Backbones}

In the main paper we performed experiments with a ResNet-18~\cite{he2016deep}, but in Table \ref{tab:effb0_results} we report performance with an EfficientNet-B0 backbone~\cite{efficientnet} to match the backbone used by prior work. Since the difference in performance between the two backbones is negligible, we can conclude that our performance gains cannot be attributed to a difference in the network backbone architecture. 

\begin{table*}[hbt!]
    \centering
    \setlength{\tabcolsep}{1.5pt}
    \begin{tabular}{|rl|c|ccc|c|}
    \hline
     & & Network Backbone & BBBC037 & BBBC036 & BBBC022 & Average\\
     \hline
    \textbf{(a)} & Transfer learning~\cite{efficientnet} & EfficientNet-B0 & 51.7 & 53.2 & 52.6 & 52.5 \\
        & Treatment classification~\cite{Caicedo_2018_CVPR} & EfficientNet-B0 & 52.7 & 58.6 & 56.6 & 56.0 \\
        \hline
    \textbf{(b)} 
        & TEAMs  & EfficientNet-B0 & \textbf{60.5} & 63.2 & 61.8 & 61.8 \\
        & TEAMs & ResNet-18 & \textbf{60.5} & \textbf{64.7} & \textbf{62.6} & \textbf{62.6}\\
        \hline
    \end{tabular}
    \caption{\textbf{Single cell results with different backbone CNN.}  We compare how often a pair of images that share a mechanistic class are predicted as more similar than a pair of images that share no mechanistic class. \textbf{(a)} reports adaptations of prior work, and \textbf{(b)} contains our model with different backbone CNN. The results show a negligible difference between these two backbone architectures.}
    \label{tab:effb0_results}
\end{table*}

\section{Varying Inference Experts}
To validate our choice of using an average of each expert in our Mixture-of-Experts (MoE) approach at inference we provide two points of comparison: 1. Random Expert. When computing similarity between two cell images in a pair, we select a single expert at random. 2. Oracle Expert. We use the ground truth expert at test time, which is typically not known, but it can provide a comparison of the supposedly ``optimal'' choice. 
 
 Table~\ref{tab:experts_main_results} reports the performance of the different methods of using our trained experts.  We first note that both the oracle and averaging approach used by our TEAMs model significantly outperform selecting an expert to use at random. Although the oracle expert has the highest accuracy for the image triplets considering only treatments, it performs significantly worse when separating from controls, making the averaging approach overall better when summing the two scores (\ie, TEAMs = 62.6 + 72.5 = 135.1 vs.\ Oracle Expert = 64.0 + 69.3 = 133.3.) This result validates our hypothesis that aggregating across experts can minimize the bias due to technical variations that are prevalent when using just a single expert on our task.  We note that a similar observation was made by Wang~\etal~\cite{wang2020fair} when creating image classification models that were fairer predictions across common subgroups in a dataset.

\begin{table*}[hbt!]
    \centering
    \setlength{\tabcolsep}{1.5pt}
    \begin{tabular}{|l|ccc|c||ccc|c|}
    \hline
    & \multicolumn{4}{|c||}{separate cell treatments} & \multicolumn{4}{|c|}{separate from controls}\\
    \cline{2-9}
     & BBBC037 & BBBC036 & BBBC022 & Average & BBBC037 & BBBC036 & BBBC022 & Average\\
     \hline
        Random Expert & 58.9 & 60.3 & 58.8 & 59.3 & 61.7 & 75.6 & 67.8 & 68.4\\
        Oracle Expert & \textbf{62.1} & \textbf{66.1} & \textbf{63.7} & \textbf{64.0} & \textbf{64.5} & 73.8 & 69.6 & 69.3\\
        Average Expert (TEAMs) & 60.5 & 64.7 & 62.6 & 62.6 & 63.6 & \textbf{81.5} & \textbf{72.4} & \textbf{72.5}\\
        \hline
    \end{tabular}
    \caption{\textbf{Single cell results with different methods of selecting an expert.} We note that oracle expert overfits to the training data, improving performance on separating by treatment slightly, but significantly hurting performance on separating from controls.  This suggests that by aggregating all the experts we are able to avoid some bias due to technical variation.}
    \label{tab:experts_main_results}
\end{table*}

\section{Results on Different Splits of Data}
We perform 5-fold cross-validation on each pre-processed Cell Painting dataset (BBBC037, BBBC036, BBBC022). The results for each split can be found in Table~\ref{tab:T-teams} for BBBC037,  Table~\ref{tab:C-teams} for BBBC036, and  Table~\ref{tab:B-teams} for BBBC022. The results show that TEAMs not only has much higher accuracy scores but also reduces variance across splits for two of the three datasets, with only a minor increase in the variance on the third dataset.  This demonstrates how TEAMs provides more stability in the results across splits in addition to higher performance.
\begin{table*}
    \centering
    \setlength{\tabcolsep}{1.5pt}
    \begin{tabular}{|rl|ccccc|c|c|}
    \hline
    & & \multicolumn{7}{|c|}{Single Cell Results}\\
    \cline{3-9}
     & & Split 1 & Split 2 & Split 3 & Split 4 & Split 5 & Average & Standard Deviation\\
     \hline
        & Transfer learning~\cite{efficientnet} & 51.9 & 51.8 & 51.6 & 51.8 & 51.5& 51.7& 0.16\\
        & Treatment classification~\cite{Caicedo_2018_CVPR} & 56.4 & 57.2 & 56.3 & 56.2 & 57.0 & 56.6 & 0.45 \\
        & TEAMs & 60.0 & 61.6 & 59.8 & 60.0 & 60.9 & 60.5& 0.77\\
        \hline
      & & \multicolumn{7}{|c|}{Treatment-Level Results}\\
      \cline{3-9}
    & & Split 1 & Split 2 & Split 3 & Split 4 & Split 5 & Average & Standard Deviation\\
      \hline
        & Transfer learning~\cite{efficientnet} & 68.5 & 72.6 & 69.4 & 67.2 & 73.3 & 70.2 & 2.64\\
        & Treatment classification~\cite{Caicedo_2018_CVPR} & 76.3 & 86.4 & 87.2 & 77.0 & 79.2 & 81.2 & 5.21\\
        & TEAMs & 83.0 & 85.1 & 85.6 & 86.2 & 84.1& 84.8& 1.27\\
        \hline
    \end{tabular}
    \caption{\textbf{Results on 5 splits of BBBC037 dataset.}}
    \label{tab:T-teams}
\end{table*}
\begin{table*}
    \centering
    \setlength{\tabcolsep}{1.5pt}
    \begin{tabular}{|rl|ccccc|c|c|}
    \hline
    & & \multicolumn{7}{|c|}{Single Cell Results}\\
    \cline{3-9}
     & & Split 1 & Split 2 & Split 3 & Split 4 & Split 5 & Average & Standard Deviation\\
     \hline
        & Transfer learning~\cite{efficientnet} & 52.3 & 53.8 & 53.4 & 54.0 & 52.4 & 53.2 & 0.79\\
        & Treatment classification~\cite{Caicedo_2018_CVPR} & 51.7 & 52.9 & 53.4 & 53.5 & 52.0 & 52.7 & 0.82\\
        & TEAMs & 63.5 & 66.6 & 64.4 & 65.7 & 63.2& 64.7& 1.45\\
        \hline
      & & \multicolumn{7}{|c|}{Treatment-Level Results}\\
      \cline{3-9}
    & & Split 1 & Split 2 & Split 3 & Split 4 & Split 5 & Average & Standard Deviation\\
      \hline
        & Transfer learning~\cite{efficientnet} & 57.5 & 60.8 & 59.8 & 59.0 & 57.6 & 58.9 & 1.42\\
        & Treatment classification~\cite{Caicedo_2018_CVPR} & 55.0 & 55.9 & 55.6 & 56.5 & 55.2 & 55.6 & 0.59\\
        & TEAMs & 65.4 & 71.3 & 66.0 & 67.9 & 66.6 & 67.4 & 2.35\\
        \hline
    \end{tabular}
    \caption{\textbf{Results on 5 splits of BBBC036 dataset.}  }
    \label{tab:C-teams}
\end{table*}

\begin{table*}
    \centering
    \setlength{\tabcolsep}{1.5pt}
    \begin{tabular}{|rl|ccccc|c|c|}
    \hline
    & & \multicolumn{7}{|c|}{Single Cell Results}\\
    \cline{3-9}
     & & Split 1 & Split 2 & Split 3 & Split 4 & Split 5 & Average & Standard Deviation\\
     \hline
        & Transfer learning~\cite{efficientnet} & 52.0 & 52.8 & 51.8 & 53.0 & 53.3 & 52.6 & 0.65\\
        & Treatment classification~\cite{Caicedo_2018_CVPR} & 58.1 & 59.0 & 57.9 & 59.1 & 58.7 & 58.6 & 0.54\\
        & TEAMs & 62.5 & 62.1 & 62.3 & 63.0 & 62.9& 62.6 & 0.39\\
        \hline
      & & \multicolumn{7}{|c|}{Treatment-Level Results}\\
      \cline{3-9}
    & & Split 1 & Split 2 & Split 3 & Split 4 & Split 5 & Average & Standard Deviation\\
      \hline
        & Transfer learning~\cite{efficientnet} & 54.1 & 54.5 & 52.4 & 55.2 & 55.2 & 54.3 & 1.15\\
        & Treatment classification~\cite{Caicedo_2018_CVPR} & 62.2 & 63.2 & 62.7 & 65.3 & 63.2 & 63.3 & 1.18\\
        & TEAMs & 63.9 & 63.7 & 63.7 & 65.0 & 64.7 & 64.2& 0.61\\
        \hline
    \end{tabular}
    \caption{\textbf{Results on 5 splits of BBBC022 dataset.}  }
    \label{tab:B-teams}
\end{table*}

\end{document}